\documentclass{article}

\usepackage{microtype}
\usepackage{graphicx}
\usepackage{subfigure}
\usepackage{booktabs} 
\usepackage{hyperref}

\usepackage{amsmath}
\usepackage{amsfonts}

\usepackage{hyperref}

\usepackage[accepted]{icml2018}
\icmltitlerunning{Learning Dynamics of Linear Denoising Autoencoders}

\newcommand{\customfootnotetext}[2]{{
  \renewcommand{\thefootnote}{#1}
  \footnotetext[0]{#2}}}

\begin{document}

\twocolumn[
\icmltitle{Learning Dynamics of Linear Denoising Autoencoders}




\begin{icmlauthorlist}
\icmlauthor{Arnu Pretorius}{cs,cair}
\icmlauthor{Steve Kroon}{cs,cair}
\icmlauthor{Herman Kamper}{ee}
\end{icmlauthorlist}

\icmlaffiliation{cs}{Computer Science Division, Stellenbosch University, South Africa}
\icmlaffiliation{cair}{CSIR/SU Centre for Artificial Intelligence Research, }
\icmlaffiliation{ee}{Department of Electrical and Electronic Engineering, Stellenbosch University, South Africa}

\icmlcorrespondingauthor{Steve Kroon}{kroon@sun.ac.za}

\icmlkeywords{Machine Learning, ICML}

\vskip 0.3in
]



\printAffiliationsAndNotice{} 

\begin{abstract}
Denoising autoencoders (DAEs) have proven useful for unsupervised representation learning, but a thorough theoretical understanding is still lacking of how the input noise influences learning. Here we develop theory for how noise influences learning in DAEs. By focusing on linear DAEs, we are able to derive analytic expressions that exactly describe their learning dynamics.
We verify our theoretical predictions with simulations as well as experiments on MNIST and \mbox{CIFAR-10}. The theory illustrates how, when tuned correctly, noise allows DAEs to ignore low variance directions in the inputs while learning to reconstruct them. Furthermore, in a comparison of the learning dynamics of DAEs to standard regularised autoencoders, we show that noise has a similar regularisation effect to weight decay, but with faster training dynamics.
We also show that our theoretical predictions approximate learning dynamics on real-world data and qualitatively match observed dynamics in nonlinear DAEs.$^\textrm{*}$ 
\end{abstract}
\section{Introduction}
\label{sec:intro}

The goal of unsupervised learning is to uncover hidden structure in unlabelled data, often in the form of latent feature representations.\customfootnotetext{}{}\customfootnotetext{*}{Code to reproduce all the results in this paper is available at: \\ \href{https://github.com/arnupretorius/lindaedynamics_icml2018}{https://github.com/arnupretorius/lindaedynamics\_icml2018}}
One popular type of model, an autoencoder, does this by trying to reconstruct its input~\cite{bengio2007greedy}.
Autoencoders have been used in various forms to address problems in machine translation~\cite{ap2014autoencoder,tu2017neural}, speech processing~\cite{elman+zipser_jasa88,zeiler+etal_icassp13}, and computer vision~\cite{rifai2011contractive,larsson2017discovery}, to name just a few areas. Denoising autoencoders (DAEs) are an extension of autoencoders which learn latent features by reconstructing data from \textit{corrupted} versions of the inputs~\cite{vincent2008extracting}. Although this corruption step typically leads to improved performance over standard autoencoders, a theoretical understanding of its effects remains incomplete. In this paper, we provide new insights into the inner workings of DAEs by analysing the learning dynamics of linear DAEs.

We specifically build on the work of \citet{saxe2013exact,saxe2013learning}, who studied the learning dynamics of deep linear networks in a supervised regression setting. By analysing the gradient descent weight update steps as time-dependent differential equations (in the limit as the learning rate approaches a small value), \citet{saxe2013exact} were able to derive exact solutions for the learning trajectory of these networks as a function of training time.
Here we  extend their approach to linear DAEs. 
To do this, we use the expected reconstruction loss over the noise distribution as an objective (requiring a different decomposition of the input covariance) as a tractable way to incorporate noise into our analytic solutions.
This approach yields exact equations which can predict the learning trajectory of a linear DAE. 


Our work here shares the motivation of many recent studies~\cite{advani2017high, pennington2017nonlinear,pennington2017geometry,nguyen2017loss,dinh2017sharp,louart2017random,swirszcz2017local,lin2016does,neyshabur2017geometry,
soudry2017exponentially,pennington2017resurrecting} working towards a better theoretical understanding of neural networks and their behaviour. Although we focus here on a theory for \textit{linear} networks, such networks have learning dynamics that are in fact \textit{nonlinear}. 
Furthermore, analyses of linear networks have also proven useful in understanding the behaviour of nonlinear neural networks~\cite{saxe2013exact,advani2017high}.

First we introduce linear DAEs (\S\ref{sec:ldae}). We then derive analytic expressions for their nonlinear learning dynamics (\S\ref{sec:dynamics}), and verify our solutions in simulations (\S\ref{sec:noise}) which show how noise can influence the shape of the loss surface and change the rate of convergence for gradient descent optimisation. We also find that an appropriate amount of noise can help DAEs ignore low variance directions in the input while learning the reconstruction mapping.
In the remainder of the paper, we compare DAEs to standard regularised autoencoders and show that our theoretical predictions match both simulations (\S\ref{sec:simulations}) and experimental results on MNIST and CIFAR-10 (\S\ref{sec:exp}).
We specifically find that while the noise in a DAE has an equivalent effect to standard weight decay, the DAE exhibits faster learning dynamics. 
We also show that our observations hold qualitatively for nonlinear DAEs.
\section{Linear Denoising Autoencoders}
\label{sec:ldae}


We first give the background of linear DAEs. Given training data consisting of pairs $\{(\tilde{\mathbf{x}}_i, \mathbf{x}_i), i=1, ... , N\}$, where $\tilde{\mathbf{x}}$ represents a corrupted version of the training data $\mathbf{x} \in \mathbb{R}^D$, the reconstruction loss for a single hidden layer DAE with activation function $\phi$ is given by
\begin{align*}
  \mathcal{L} = \frac{1}{2N}\sum^N_{i=1}||\mathbf{x}_i - W_2\phi(W_1\tilde{\mathbf{x}}_i)||^2.
\end{align*}
Here, $W_1 \in \mathbb{R}^{H \times D}$ and $W_2 \in \mathbb{R}^{D \times H}$ are the weights of the network with hidden dimensionality $H$. The learned feature representations correspond to the latent variable $\mathbf{z} = \phi(W_1\tilde{\mathbf{x}})$.

To corrupt an input $\mathbf{x}$, we sample a noise vector $\epsilon$, where each component is drawn i.i.d.\ from a pre-specified noise distribution with mean zero and variance $s^2$. We define the corrupted version of the input as $\tilde{\mathbf{x}} = \mathbf{x} + \epsilon$. This ensures that the expectation over the noise remains unbiased, i.e.\ $\mathbb{E}_\epsilon(\tilde{\mathbf{x}}) = \mathbf{x}.$ 


Restricting our scope to linear neural networks, with $\phi(a) = a$, the loss in expectation over the noise distribution~is 
\begin{align}
   \mathbb{E}_{\epsilon}\left [\mathcal{L} \right ] & = \frac{1}{2N} \sum^N_{i=1}||\mathbf{x}_i - W_2W_1\mathbf{x}_i||^2 \nonumber \\
  & \textcolor{white}{white ce} + \frac{s^2}{2}\text{tr}(W_2W_1W_1^TW_2^T),
  \label{eq: dae loss}
\end{align}
See the supplementary material for the full derivation.
\section{Learning Dynamics of Linear DAEs}
\label{sec:dynamics}

Here we derive the learning dynamics of linear DAEs, beginning with a brief outline to build some intuition.

The weight update equations for a linear DAE can be formulated as time-dependent differential equations in the limit as the gradient descent learning rate becomes small~\cite{saxe2013exact}. The task of an ordinary (undercomplete) linear autoencoder is to learn the identity mapping that reconstructs the original input data. The matrix corresponding to this learned map will essentially be an  approximation of the full identity matrix that is of rank equal to the input dimension. It turns out that tracking the temporal updates of this mapping represents a difficult problem that involves dealing with coupled differential equations, since both the on-diagonal and off-diagonal elements of the weight matrices need to be considered in the approximation dynamics at each time step. 

To circumvent this issue and make the analysis tractable, we follow the methodology introduced in \citet{saxe2013exact}, which is to: $(1)$ decompose the input covariance matrix using an eigenvalue decomposition; $(2)$ rotate the weight matrices to align with these computed directions of variation; and $(3)$~use an orthogonal initialisation strategy to diagonalise the composite weight matrix $W = W_2W_1$. The important difference in our setting, is that additional constraints are brought about through the injection of noise. 

The remainder of this section outlines this derivation for the exact solutions to the learning dynamics of linear DAEs.

\subsection{Gradient descent update}
\label{sec:gd}

Consider a continuous time limit approach to studying the learning dynamics of linear DAEs. This is achieved by choosing a sufficiently small learning rate $\alpha$ for optimising the loss in \eqref{eq: dae loss} using gradient descent. The update for $W_1$ in a single gradient descent step then takes the form of a time-dependent differential equation 
\begin{align*}
  \tau\frac{d}{dt}W_1 & = \sum^N_{i=1} W_2^T \left ( \mathbf{x}_i\mathbf{x}_i^T - W_2W_1\mathbf{x}_i\mathbf{x}_i^T \right ) \\ 
  & \textcolor{white}{white sp} - \varepsilon W_2^TW_2W_1 \\
  & = W_2^T \left ( \Sigma_{xx} - W_2W_1\Sigma_{xx} \right ) -  \varepsilon W_2^TW_2W_1.
\end{align*}
Here $t$ is the time measured in epochs, $\tau = \frac{N}{\alpha}, \varepsilon = N s^2$ and $\Sigma_{xx} = \sum^N_{i=1}\mathbf{x}_i\mathbf{x}_i^T$, represents the input covariance matrix. Let the eigenvalue decomposition of the input covariance be $\Sigma_{xx} = V \Lambda V^T$, where $V$ is an orthogonal matrix and denote the eigenvalues $\lambda_j = [\Lambda]_{jj}$, with $\lambda_1 \geq \lambda_2 \geq \dots \geq \lambda_D$. The update can then be rewritten as
\begin{align*}
  \tau\frac{d}{dt}W_1 
  & = W_2^TV \left (\Lambda - V^TW_2W_1V\Lambda \right)V^T \\
  & \textcolor{white}{more white space} - \varepsilon W_2^TW_2W_1.
\end{align*}
The weight matrices can be rotated to align with the directions of variation in the input by performing the rotations $\overline{W}_1 = W_1V$ and $\overline{W}_2 = V^TW_2$. Following a similar derivation for $W_2$, the weight updates become
\begin{align*}
  \tau\frac{d}{dt}\overline{W}_1 & = \overline{W}_2^T \left (\Lambda - \overline{W}_2\overline{W}_1\Lambda \right) - \varepsilon \overline{W}_2^T\overline{W}_2\overline{W}_1 \\
  \tau\frac{d}{dt}\overline{W}_2 & = \left (\Lambda - \overline{W}_2\overline{W}_1\Lambda \right)\overline{W}_1^T - \varepsilon \overline{W}_2\overline{W}_1\overline{W}^T_1.
\end{align*}  

\subsection{Orthogonal initialisation and scalar dynamics} \label{subsec: ortho initialisation}

To decouple the dynamics, we can set $W_2 = VD_2R^T$ and $W_1 = RD_1V^T$, where $R$ is an arbitrary orthogonal matrix and $D_2$ and $D_1$ are diagonal matrices. This results in the product of the realigned weight matrices
\begin{align*}
	\overline{W}_2\overline{W}_1 = V^TVD_2R^TRD_1V^TV = D_2D_1
\end{align*}
to become diagonal. 
The updates now reduce to the following scalar dynamics that apply independently to each pair of diagonal elements $w_{1j}$ and $w_{2j}$ of $D_1$ and $D_2$ respectively:
\begin{align}
  \tau\frac{d}{dt}w_{1j} & = w_{2j}  \lambda_j \left (1 - w_{2j}w_{1j} \right) - \varepsilon w^2_{2j}w_{1j} \label{eq:dw1} \\ 
  \tau\frac{d}{dt}w_{2j} & = w_{1j}  \lambda_j \left (1 - w_{2j}w_{1j} \right) - \varepsilon w_{2j}w^2_{1j}. \label{eq:dw2}
\end{align}
Note that the same dynamics stem from gradient descent on the loss given by
\begin{align}
  \ell = \sum^D_{j=1}\frac{\lambda_j}{2\tau}(1 - w_{2j}w_{1j})^2 + \sum^D_{j=1}\frac{\varepsilon}{2\tau}(w_{2j}w_{1j})^2.
  \label{eq: scalar loss}
\end{align}
By examining \eqref{eq: scalar loss}, it is evident that the degree to which the first term will be reduced will depend on the magnitude of the associated eigenvalue $\lambda_j$. However, for directions in the input covariance $\Sigma_{xx}$ with relatively little variation the decrease in the loss from learning the identity map will be negligible and is likely to result in overfitting (since little to no signal is being captured by these eigenvalues). The second term in~\eqref{eq: scalar loss} is the result of the input corruption and acts as a suppressant on the magnitude of the weights in the learned mapping. Our interest is to better understand the interplay between these two terms during learning by studying their scalar learning dynamics. 

\subsection{Exact solutions to the dynamics of learning} \label{subsec: Exact solutions for the dynamics of learning}
\label{sec:exact_solutions}

As noted above, the dynamics of learning are dictated by the value of $w = w_2w_1$ over time.
An expression can be derived for $w(t)$ by using a hyperbolic change of coordinates in \eqref{eq:dw1} and \eqref{eq:dw2}, letting $\theta$ parameterise points along a dynamics trajectory represented by the conserved quantity $w_2^2 - w_1^2 = \pm c_0$. This relies on the fact that $\ell$ is invariant under a scaling of the weights such that $w = (w_1/c)(cw_2) = w_2w_1$ for any constant $c$ \citep{saxe2013exact}.   
Starting at any initial point $(w_1, w_2)$ the dynamics are
\begin{align}
w(t) = \frac{c_0}{2}\text{sinh}\left (\theta_t \right),
\label{eq: exact solution}
\end{align}
with
\begin{align*}
    \theta_t = 2\text{tanh}^{-1}\left [ \frac{(1-E)\left (\zeta^2 - \beta^2 - 2\beta\delta \right) - 2(1 + E)\zeta \delta}{(1-E)\left ( 2\beta + 4\delta\right) - 2(1+E)\zeta} \right ]
\end{align*}
where \mbox{$\beta = c_0\left(1 + \frac{\varepsilon}{\lambda}\right)$}, $\zeta = \sqrt{\beta^2 + 4}$, $\delta = \text{tanh}\left(\frac{\theta_0}{2}\right)$ and $E = e^{\zeta\lambda t/\tau}.$ Here $\theta_0$ depends on the initial weights $w_1$ and $w_2$ through the relationship \mbox{$\theta_0 = \text{sinh}^{-1}(2w/c_0)$}.
The derivation for $\theta_t$ involves rewriting $\tau \frac{d}{dt}w$ in terms of $\theta$, integrating over the interval $\theta_0$ to $\theta_t$, and finally rearranging terms to get an expression for $\theta(t) \equiv \theta_t$ (see the supplementary material for full details). To derive the learning dynamics for different noise distributions, the corresponding $\varepsilon$ must be computed and used to determine $\beta$ and~$\zeta$. For example, sampling noise from a Gaussian distribution such that $\epsilon \sim \mathcal{N}(\mathbf{0}, \sigma^2I)$, gives $\varepsilon = N\sigma^2$. Alternatively, if $\epsilon$ is distributed according to a zero-mean Laplace distribution with scale parameter $b$, then $\varepsilon = 2Nb^2$. 


\section{The Effects of Noise: a Simulation Study}
\label{sec:noise}

\begin{figure*}[h]
   \centering
  \includegraphics[width=0.95\linewidth]{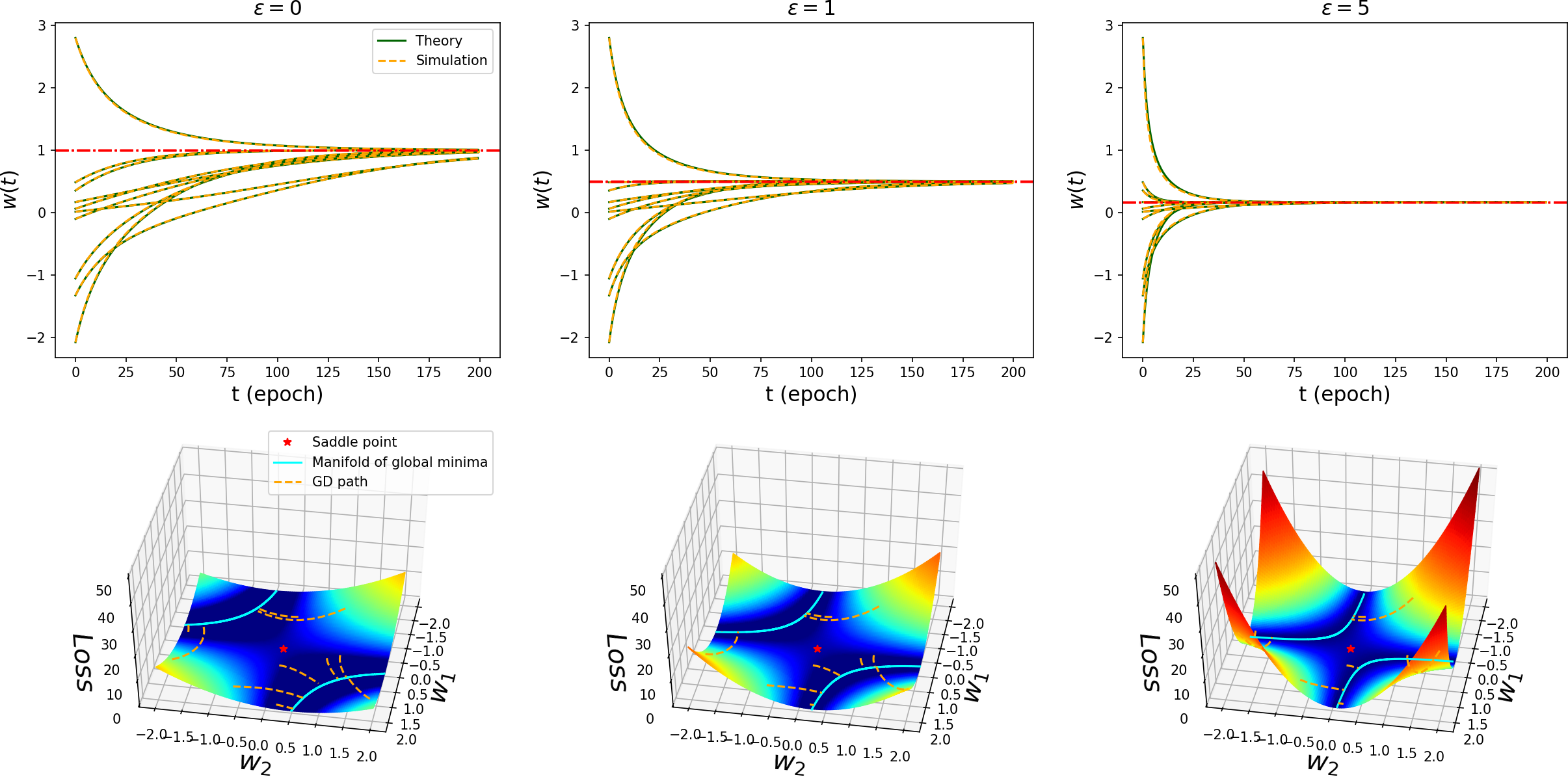}
   \caption{\emph{Learning dynamics, loss surface and gradient descent paths for linear denoising autoencoders.} \textbf{Top}: Learning dynamics for each simulated run (dashed orange lines) together with the theoretically predicted learning dynamics (solid green lines). The red line in each plot indicates the final value of the resulting fixed point solution $w^*$. \textbf{Bottom}: The loss surface corresponding to the loss $\ell_\lambda = \frac{\lambda}{2}(1 - w_2w_1)^2 + \frac{\varepsilon}{2}(w_2w_1)^2$ for $\lambda = 1$, as well as the gradient descent paths (dashed orange lines) for randomly initialised weights. The cyan hyperbolas represent the global minimum loss manifold that corresponds to all possible combinations of $w_2$ and $w_1$ that minimise $\ell_\lambda$. \textbf{Left}: $\varepsilon = 0, w^* = 1$. \textbf{Middle}: $\varepsilon = 1, w^* = 0.5$. \textbf{Right}: $\varepsilon = 5, w^* = 1/6$.}
   \label{fig: effect of noise}
\end{figure*}

Since {the expression for the learning dynamics of a linear DAE} in~\eqref{eq: exact solution} evolve independently for each direction of variation in the input, it is enough to study the effect that noise has on learning for a single eigenvalue $\lambda$. To do this we trained a scalar linear DAE to minimise the loss $\ell_\lambda = \frac{\lambda}{2}(1 - w_2w_1)^2 + \frac{\varepsilon}{2}(w_2w_1)^2$ with $\lambda = 1$ using gradient descent.
Starting from several different randomly initialised weights $w_1$ and $w_2$, we compare the simulated dynamics with those predicted by equation~\eqref{eq: exact solution}. The top row in Figure~\ref{fig: effect of noise} shows the exact fit between the predictions and numerical simulations for different noise levels, $\varepsilon = 0, 1, 5$. 

The trajectories in the top row of Figure \ref{fig: effect of noise} converge to the optimal solution at different rates depending on the amount of injected noise. Specifically, adding more noise results in faster convergence. However, the trade-off in~\eqref{eq: scalar loss} ensures that the fixed point solution also diminishes in magnitude. 


To gain further insight, we also visualise the associated loss surfaces for each experiment in the bottom row of Figure~\ref{fig: effect of noise}. Note that even though the scalar product $w_2w_1$ defines a linear mapping, the minimisation of $\ell_\lambda$ with respect to $w_1$ and $w_2$ is a non-convex optimisation problem. The loss surfaces in Figure~\ref{fig: effect of noise} each have an unstable saddle point at $w_2=w_1=0$ (red star) with all remaining fixed points lying on a minimum loss manifold (cyan curve). This manifold corresponds to the different possible combinations of $w_2$ and $w_1$ that minimise $\ell_\lambda$. The paths that gradient descent follow from various initial starting weights down to points situated on the manifold are represented by dashed orange~lines. 

For a fixed value of $\lambda$, adding noise warps the loss surface making steeper slopes and pulling the minimum loss manifold in towards the saddle point. Therefore, steeper descent directions cause learning to converge at a faster rate to fixed points that are smaller in magnitude. This is the result of a sharper curving loss surface and the minimum loss manifold lying closer to the origin. 


We can compute the fixed point solution for any pair of initial starting weights (not on the saddle point) by taking the derivative
\begin{align*}
	\frac{d \ell_\lambda}{d w} = -\frac{\lambda}{\tau}(1 - w) + \frac{\varepsilon}{\tau}w,
\end{align*}
and setting it equal to zero to find $w^* = \frac{\lambda}{\lambda + \varepsilon}$. This solution reveals the interaction between the input variance associated with $\lambda$ and the noise $\varepsilon$. For large eigenvalues for which $\lambda \gg \varepsilon$, the fixed point will remain relatively unaffected by adding noise, i.e., $w^* \approx 1$. In contrast, if $\lambda \ll \varepsilon$, the noise will result in $w^* \approx 0$. This means that over a distribution of eigenvalues, an appropriate amount of noise can help a DAE to ignore low variance directions in the input data while learning the reconstruction. In a practical setting, this motivates the tuning of noise levels on a development set to prevent overfitting.
\section{The Relationship Between Noise and Weight Decay}
\label{sec:simulations}

It is well known that adding noise to the inputs of a neural network is equivalent to a form of regularisation~\cite{bishop1995training}. Therefore, to further understand the role of noise in linear DAEs we compare the dynamics of noise to those of explicit regularisation in the form of \textit{weight decay}~\cite{krogh1992simple}. The reconstruction loss for a linear weight decayed autoencoder (WDAE) is given~by
\begin{align}
	\frac{1}{2N}\sum^N_{i=1}||\mathbf{x}_i - W_2W_1\mathbf{x}_i||^2 + \frac{\gamma}{2} \left (||W_1||^2 + ||W_2||^2 \right)
\label{eq: dae loss plus reg}
\end{align}
where $\gamma$ is the penalty parameter that controls the amount of regularisation applied during learning. Provided that the weights of the network are initialised to be small, it is also possible (see supplementary material) to derive scalar dynamics of learning from~\eqref{eq: dae loss plus reg} as
\begin{align}
w_\gamma(t) = \frac{\xi E_\gamma}{E_\gamma - 1 + \xi/w_0},
\label{eq: learning dynamics final}
\end{align}
where $\xi = (1-N\gamma/\lambda)$ and $E_\gamma = e^{2\xi t/\tau}$. 

\begin{figure*}[!h]
   \centering
    \includegraphics[width=0.95\linewidth]{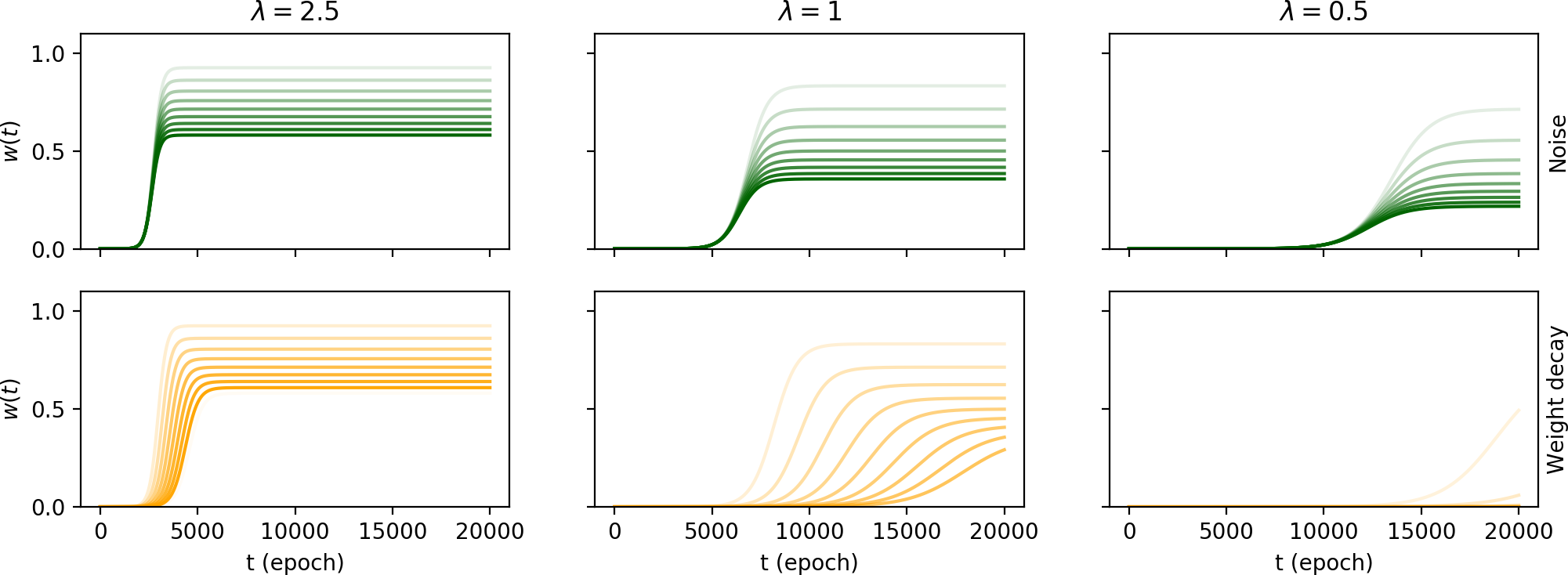}
   \caption{\textit{Theoretically predicted learning dynamics for noise compared to weight decay for linear autoencoders.} \textbf{Top}: Noise dynamics (green), darker line colours correspond to larger amounts of added noise. \textbf{Bottom}: Weight decay dynamics (orange), darker line colours correspond to larger amounts of regularisation. \textbf{Left to right}: Eigenvalues $\lambda = 2.5, 1$ and $0.5$ associated with high to low variance.}
   \label{fig: noise vs reg learning dynamics}
\end{figure*}

\begin{figure*}[!htb]
   \centering
    \includegraphics[width=0.95\linewidth]{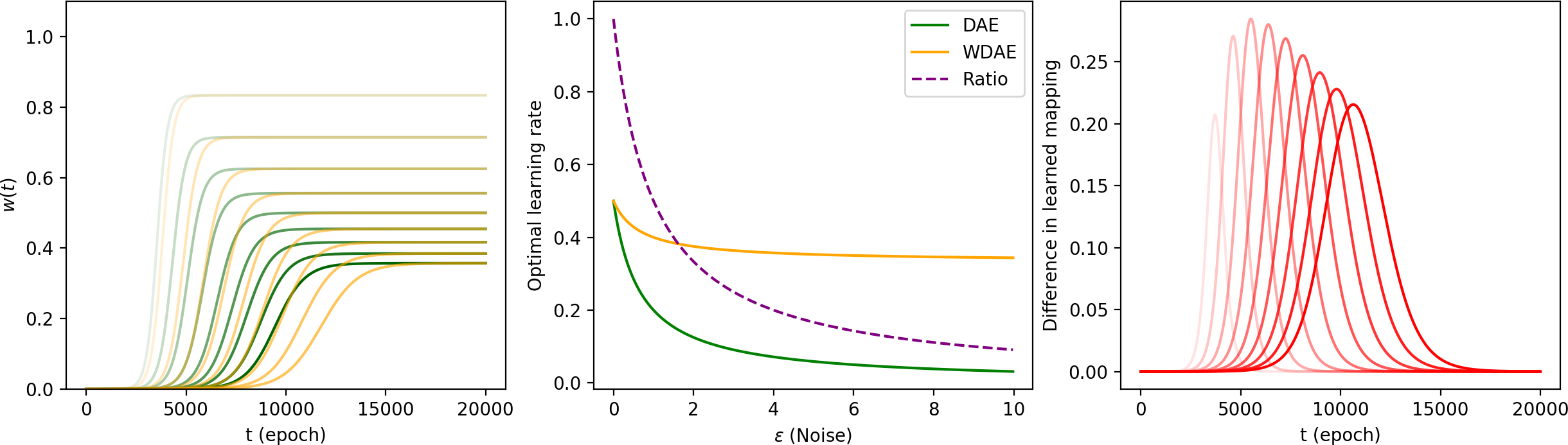}
   \caption{\textit{Learning dynamics for optimal discrete time learning rates ($\lambda = 1$).} \textbf{Left}: Dynamics of DAEs (green) vs. WDAEs (orange), where darker line colours correspond to larger amounts noise or weigh decay. \textbf{Middle}: Optimal learning rate as a function of noise $\varepsilon$ for DAEs, and for WDAEs using an equivalent amount of regularisation $\gamma = \lambda\varepsilon/(\lambda + \varepsilon)$.  \textbf{Right}: Difference in mapping over time.}
   \label{fig: optimal learning rates}
\end{figure*}

\begin{figure*}[!h]
   \centering
  \includegraphics[width=0.95\linewidth]{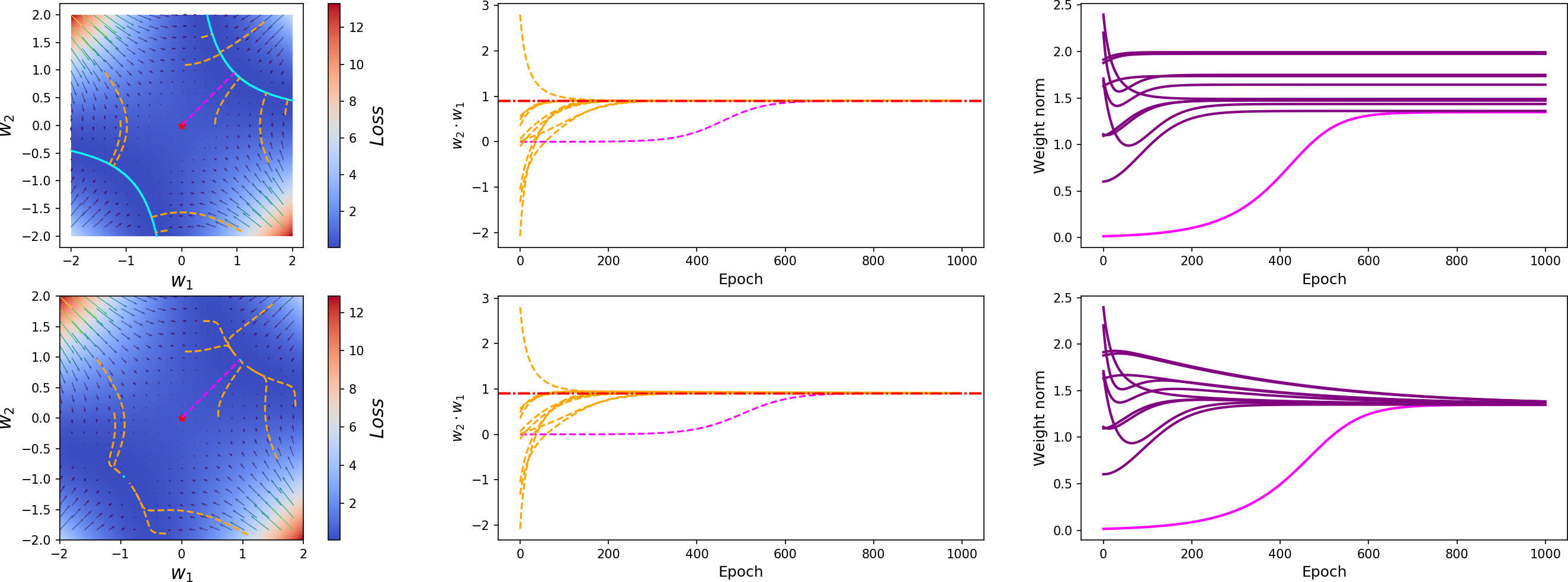}
   \caption{\emph{The effect of noise versus weight decay on the norm of the weights during learning.} \textbf{Left}: Two-dimensional loss surface $\ell_\lambda = \frac{\lambda}{2}(1 - w_2w_1)^2 + \frac{\varepsilon}{2}(w_2w_1)^2 + \frac{\gamma}{2}(w^2_2+w^2_1)$. Gradient descent paths (orange/magenta dashed lines), minimum loss manifold (cyan curves), saddle point (red star). \textbf{Middle}: Simulated learning dynamics. \textbf{Right}: Norm of the weights over time for each simulated run. \textbf{Top}: Noise with $\lambda = 1, \varepsilon = 0.1$ and $\gamma = 0$. \textbf{Bottom}: Weight decay with $\lambda = 1, \varepsilon = 0$ and $\gamma = \lambda (0.1)/(\lambda + 0.1) = 0.091$. The magenta line in each plot corresponds to a simulated run with small initialised weights.}
   \label{fig: noise vs reg weight norm}
\end{figure*}
Figure~\ref{fig: noise vs reg learning dynamics} compares the learning trajectories of linear DAEs and WDAEs over time (as measured in training epochs) for $\lambda = 2.5, 1, 0.5$ and $0.1$. The dynamics for both noise and weight decay exhibit a sigmoidal shape with an initial period of inactivity followed by rapid learning, finally reaching a plateau at the fixed point solution. Figure~\ref{fig: noise vs reg learning dynamics} illustrates that the learning time associated with an eigenvalue is negatively correlated with its magnitude.
Thus, the eigenvalue corresponding to the largest amount of variation explained is the quickest to escape inactivity during learning.

The colour intensity of the lines in Figure~\ref{fig: noise vs reg learning dynamics} correspond to the amount of noise or regularisation applied in each run, with darker lines indicating larger amounts. In the continuous time limit with equal learning rates, when compared with noise dynamics, weight decay experiences a delay in learning such that the initial inactive period becomes extended for every eigenvalue, whereas adding noise has no effect on learning time. In other words, starting from small weights, noise injected learning is capable of providing an equivalent regularisation mechanism to that of weight decay in terms of a constrained fixed point mapping, but with zero time delay.

However, this analysis does not take into account the practice of using well-tuned stable learning rates for discrete optimisation steps. We therefore consider 
the impact on training time when using optimised learning rates for each approach. By using second order information from the Hessian as in \citet{saxe2013exact}, (here of the expected reconstruction loss with respect to the scalar weights), we relate the optimal learning rates for linear DAEs and WDAEs, where each optimal rate is inversely related to the amount of noise/regularisation applied during training (see supplementary material). The ratio of the optimal DAE rate to that for the WDAE is
\begin{align}
	R = \frac{2\lambda + \gamma}{2\lambda + 3\varepsilon}.
	\label{eq: optimal learning rates}
\end{align}
Note that the ratio in \eqref{eq: optimal learning rates} will essentially be equal to one for eigenvalues that are significantly larger than both $\varepsilon$ and $\gamma$, with deviations from unity only manifesting for smaller values of $\lambda$. 

Furthermore, weight decay and noise injected learning result in equivalent scalar solutions when their parameters are related by $\gamma = \frac{\lambda \varepsilon}{\lambda + \varepsilon}$ (see supplementary material). This leads to the following two observations. First, it shows that adding noise during learning can be interpreted as a form of weight decay where the penalty parameter $\gamma$ adapts to each direction of variation in the data. In other words, noise essentially makes use of the statistical structure of the input data to influence the amount of shrinkage that is being applied in various directions during learning. Second, together with \eqref{eq: optimal learning rates}, we can theoretically compare the learning dynamics of DAEs and WDAEs, when both equivalent regularisation and the relative differences in optimal learning rates are taken into account.

The effects of optimal learning rates (for $\lambda = 1$), are shown in Figure \ref{fig: optimal learning rates}. DAEs still exhibit faster dynamics (left panel), even when taking into account the difference in the learning rate as a function of noise, or equivalent weight decay (middle panel). In addition, for equivalent regularisation effects, the ratio of the optimal rates $R$ can be shown to be a monotonically decreasing function of the noise level, where the rate of decay depends on the size of $\lambda$. This means that for any amount of added noise, the DAE will require a slower learning rate than that of the WDAE. Even so, a faster rate for the WDAE does not seem to compensate for its slower dynamics and the difference in learning time is also shown to grow as more noise (regularisation) is applied during training (right panel).



\subsection{Exploiting invariance in the loss function}

A primary motivation for weight decay as a regulariser is that it provides solutions with smaller weight norms, producing smoother models that have better generalisation performance. Figure~\ref{fig: noise vs reg weight norm} shows the effect of noise (top row) compared to weight decay (bottom row) on the norm of the weights during learning. Looking at the loss surface for weight decay (bottom left panel), the penalty on the size of the weights acts by shrinking the minimum loss manifold down from a long curving valley to a single point (associated with a small norm solution). Interestingly, this results in gradient descent following a trajectory towards an ``invisible'' minimum loss manifold similar to the one associated with noise. However, once on this manifold, weight decay begins to exploit invariances in the loss function to changes in the weights, so as to move along the manifold down towards smaller norm solutions. This means that even when the two approaches learn the exact same mapping over time (as shown by the learning dynamics in the middle column of Figure~\ref{fig: noise vs reg weight norm}), additional epochs will cause weight decay to further reduce the size of the weights (bottom right panel).
This happens in a stage-like manner where the optimisation first focuses on reducing the reconstruction loss by learning the optimal mapping and then reduces the regularisation loss through invariance.

\subsection{Small weight initialisation and early stopping} 

It is common practice to initialise the weights of a network with small values. In fact, this strategy has recently been theoretically shown to help, along with early stopping, to ensure good generalisation performance for neural networks in certain high-dimensional settings~\cite{advani2017high}. In our analysis however, what we find interesting about small weight initialisation is that it removes some of the differences in the learning behaviour of DAEs compared to regularised autoencoders that use weight decay. 

To see this, the magenta lines in Figure~\ref{fig: noise vs reg weight norm} show the learning dynamics for the two approaches where the weights of both the networks were initialised to small random starting values. The learning dynamics are almost identical in terms of their temporal trajectories  and have equal fixed points. However, what is interesting is the implicit regularisation that is brought about through the small initialisation. By starting small and making incremental updates to the weights, the scalar solution in both cases end up being equal to the minimum norm solution. In other words, the path that gradient descent takes from the initialisation to the minimum loss manifold, reaches the manifold where the norm of the weights happen to also be small. This means that the second phase of weight decay (where the invariance of the loss function would be exploited to reduce the regularisation penalty), is not only no longer necessary, but also does not result in a norm that is appreciably smaller than that obtained by learning with added noise. Therefore in this case, learning with explicit regularisation provides no additional benefit over that of learning with noise in terms of reducing the norm of the weights during training.


When initialising small, early stopping can also serve as a form of implicit regularisation by ensuring that the weights do not change past the point where the validation loss starts to increase~\cite{bengio2007greedy}. In the context of learning dynamics, early stopping for DAEs can be viewed as a method that effectively selects only the directions of variation deemed useful for generalisation during reconstruction, considering the remaining eigenvalues to carry no additional signal.
\section{Experimental Results}
\label{sec:exp}


To verify the dynamics of learning on real-world data sets we compared theoretical predictions with actual learning on MNIST and CIFAR-10. In our experiments we considered the following linear autoencoder networks: a regular AE, a WDAE and a DAE.

\begin{figure}[t]
   \centering
  \includegraphics[width=0.99\linewidth]{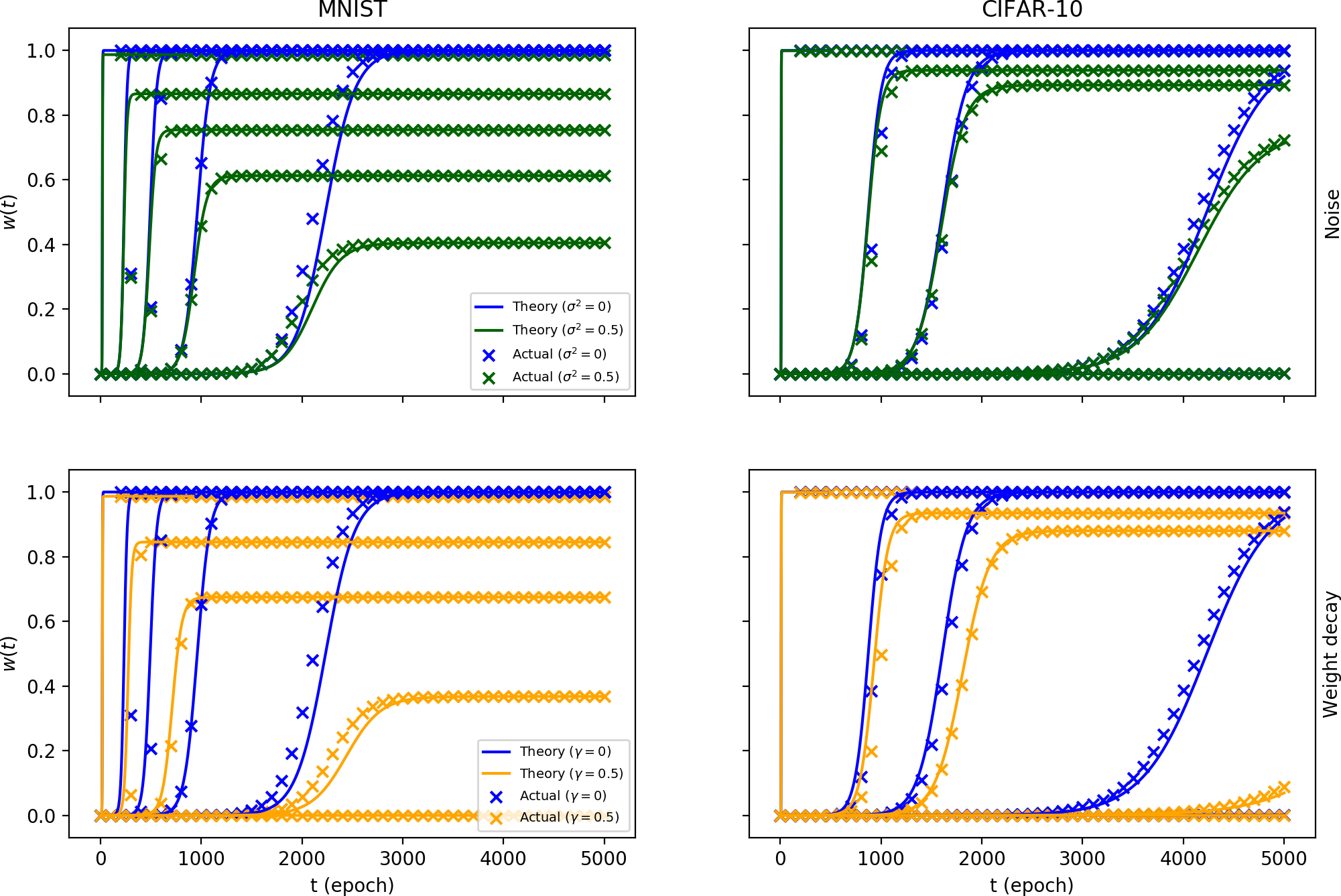} 
   \caption{\textit{Learning dynamics for MNIST and CIFAR-10.} Solid lines represent theoretical dynamics and `x' markers simulated dynamics. Shown are the mappings associated with the set of eigenvalues $\{\lambda_i, i = 1, 4, 8, 16, 32 \}$, where the remaining eigenvalues were excluded to improve readability. \textbf{Top}: Noise: AE (blue) vs. DAE with $\sigma^2 = 0.5$ (green). \textbf{Bottom}: Weight decay: AE (blue) vs. WDAE with $\gamma = 0.5$ (orange).  \textbf{Left}: MNIST. \textbf{Right}: CIFAR-10.}
   \label{fig: real-world learning dynamics}
\end{figure}

For MNIST, we trained each autoencoder with small randomly initialised weights, using $N = 50 000$ training samples for $5000$ epochs, with a learning rate $\alpha = 0.01$ and a hidden layer width of $H = 256$. For the WDAE, the penalty parameter was set at $\gamma = 0.5$ and for the DAE, $\sigma^2 = 0.5$. The results are shown in Figure~\ref{fig: real-world learning dynamics} (left column).

The theoretical predictions (solid lines) in Figure~\ref{fig: real-world learning dynamics} show good agreement with the actual learning dynamics (points). As predicted, both regularisation (orange) and noise (green) suppress the fixed point value associated with the different eigenvalues and, whereas regularisation delays learning (fewer fixed points are reached by the WDAE during training when compared to the DAE), the use of noise has no effect on training time. 

Similar agreement is shown for CIFAR-10 in the right column of Figure \ref{fig: real-world learning dynamics}. Here, we trained each network with small randomly initialised weights using $N = 30000$ training samples for $5000$ epochs, with a learning rate $\alpha = 0.001$ and a hidden dimension $H = 512$. For the WDAE, the penalty parameter was set at $\gamma = 0.5$ and for the DAE, $\sigma^2 = 0.5$.  

\begin{figure}[t]
   \centering
  \includegraphics[width=0.99\linewidth]{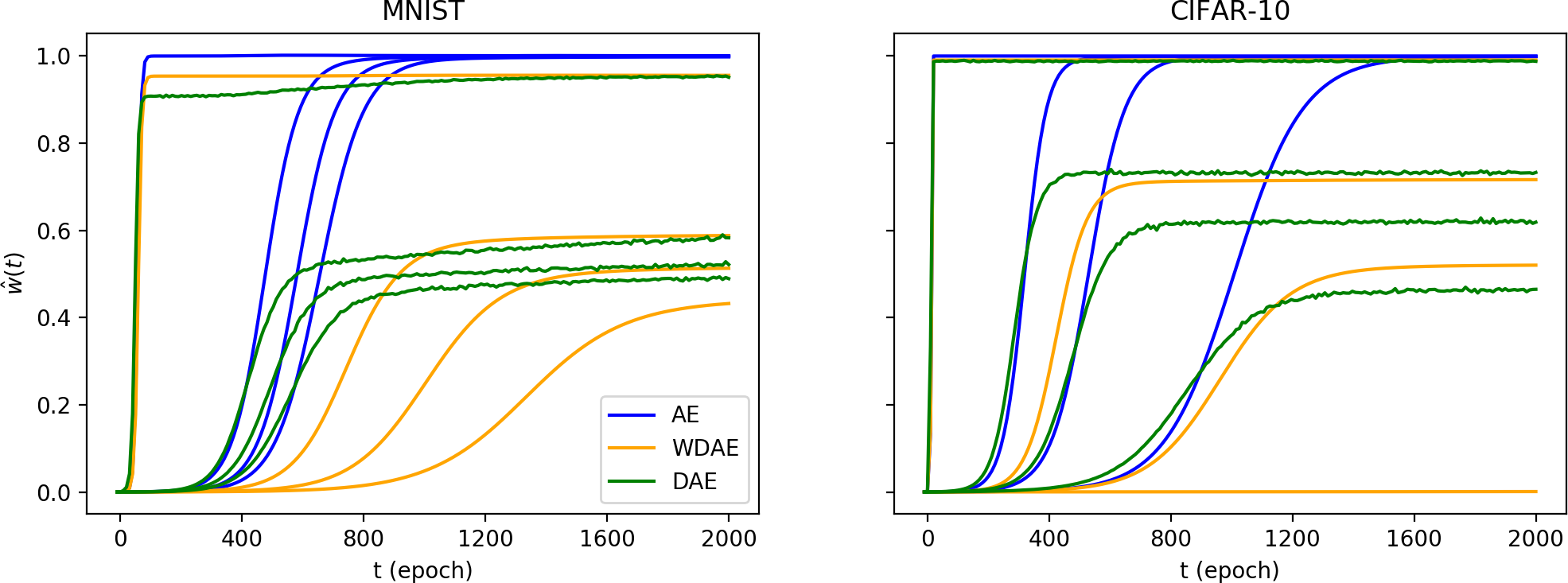} 
   \caption{\textit{Learning dynamics for nonlinear networks using ReLU activation.} AE (blue), WDAE (orange) and DAE (green). Shown are the mappings associated with the first four eigenvalues, i.e. $\{\lambda_i, i = 1, 2, 3, 4 \}$. \textbf{Left}: MNIST \textbf{Right}: CIFAR-10.}
   \label{fig: nonlinear learning dynamics}
\end{figure}

Next, we investigated whether these dynamics are at least also qualitatively present in nonlinear autoencoder networks. Figure \ref{fig: nonlinear learning dynamics} shows the dynamics of learning for nonlinear AEs, WDAEs and DAEs, using ReLU activations, trained on MNIST ($N = 50000$) and CIFAR-10 $(N=30000)$ with equal learning rates. For the DAE, the input was corrupted using sampled Gaussian noise with mean zero and $\sigma^2 = 3$. For the WDAE, the amount of weight decay was manually tuned to $\gamma = 0.0045$, to ensure that both autoencoders displayed roughly the same degree of regularisation in terms of the fixed points reached. During the course of training, the identity mapping associated with each eigenvalue  was estimated (see supplementary material), at equally spaced intervals of size $10$ epochs.

The learning dynamics are qualitatively similar to the dynamics observed in the linear case. Both noise and weight decay result in a shrinkage of the identity mapping associated with each eigenvalue. Furthermore, in terms of the number of training epochs, the DAE is seen to learn as quickly as a regular AE, whereas the WDAE incurs a delay in learning time. Although these experimental results stem from a single training run for each autoencoder, we note that wall-clock times for training may still differ because DAEs require some additional time for sampling noise. Similar results were observed when using a tanh nonlinearity and are provided in the supplementary material.
\section{Related Work}

There have been many studies aiming to provide a better theoretical understanding of DAEs. \citet{vincent2008extracting} analysed DAEs from several different perspectives, including manifold learning and information filtering, by establishing an equivalence between different criteria for learning and the original training criterion that seeks to minimise the reconstruction loss. Subsequently, \citet{vincent2011connection} showed that under a particular set of conditions, the training of DAEs can also be interpreted as a type of score matching. This connection provided a probabilistic basis for DAEs. Following this, a more in-depth analysis of DAEs as a possible generative model suitable for arbitrary loss functions and multiple types of data was given by \citet{bengio2013generalized}.

In contrast to a probabilistic understanding of DAEs, we present here an analysis of the learning process. Specifically inspired by \citet{saxe2013exact}, as well as by earlier work on supervised neural networks~\cite{opper1988learning,sanger1989optimal,baldi1989neural,saad1995exact}, we provide a theoretical investigation of the temporal behaviour of linear DAEs using derived equations that exactly describe their dynamics of learning. Specifically for the linear case, the squared error loss for the reconstruction contractive autoencoder (RCAE) introduced in \citet{alain2014regularized} is equivalent to the expected loss (over the noise) for the DAE. Therefore, the learning dynamics described in this paper also apply to linear RCAEs.

For our analysis to be tractable we used a marginalised reconstruction loss where the gradient descent dynamics are viewed in expectation over the noise distribution. Whereas our motivation is analytical in nature, marginalising the reconstruction loss tends to be more commonly motivated from the point of view of learning useful and robust feature representations at a significantly lower computational cost~\cite{chen2014marginalized,chen2015marginalizing}. This approach has also been investigated in the context of supervised learning~\cite{maaten2013learning,wang2013fast,wager2013dropout}. Also related to our work is the analysis by \citet{poole2014analyzing}, who showed that training autoencoders with noise (added at different levels of the network architecture), is closely connected to training with explicit regularisation and proposed a marginalised noise framework for noisy autoencoders.
\section{Conclusion and Future Work}

This paper analysed the learning dynamics of linear denoising autoencoders (DAEs) with the aim of providing a better understanding of the role of noise during training. By deriving exact time-dependent equations for learning, we showed how noise influences the shape of the loss surface as well as the rate of convergence to fixed point solutions. We also compared the learning behaviour of added input noise to that of weight decay, an explicit form of regularisation. We found that while the two have similar regularisation effects, the use of noise for regularisation results in faster training. We compared our theoretical predictions with actual learning dynamics on real-world data sets, observing good agreement. In addition, we also provided evidence (on both MNIST and CIFAR-10) that our predictions hold qualitatively for nonlinear DAEs. 


This work provides a solid basis for further investigation. Our analysis could be extended to nonlinear DAEs, potentially using the recent work on nonlinear random matrix theory for neural networks~\cite{pennington2017nonlinear, louart2017random}. 
Our findings indicate that appropriate noise levels help DAEs ignore low variance directions in the input; we also obtained new insights into the training time of DAEs. Therefore, future work might consider how these insights could actually be used for tuning noise levels and predicting the training time of DAEs. This would require further validation and empirical experiments, also on other datasets.
Finally, our analysis only considers the training dynamics, while a better understanding of generalisation and what influences the quality of feature representations during testing, are also of prime importance.
\section*{Acknowledgements}

We would like to thank Andrew Saxe for early discussions that got us interested in this work, as well as the reviewers for insightful comments and suggestions. We would like to thank the CSIR/SU Centre for Artificial Intelligence Research (CAIR), South Africa, for financial support. AP would also like to thank the MIH Media Lab at Stellenbosch University and Praelexis (Pty) Ltd for providing stimulating working environments for a portion of this work.

\newpage

\bibliography{ref}
\bibliographystyle{icml2018}

\clearpage

\section*{Supplementary material}

The following section provides detail omitted in the paper regarding the derivation of certain equations as well as additional comments. 

\subsection*{A. Expected loss for linear DAEs}
We derive the expected reconstruction loss over the noise distribution as presented in (1) in the paper. The expected loss can be written as
\begin{align*}
  \mathbb{E}_{\epsilon}[\mathcal{L}] = \frac{1}{2N}\sum^N_{i=1}\mathbb{E}_{\epsilon} \left [ ||\mathbf{x}_i - W_2W_1\tilde{\mathbf{x}}_i||^2 \right ].
\end{align*}
where $\tilde{\mathbf{x}}_i = \mathbf{x}_i + \epsilon_i$, with $\epsilon$ sampled from an isotropic noise distribution with component variance $s^2$. Let $SE(\tilde{\mathbf{x}}_i) = ||\mathbf{x}_i - W_2W_1\tilde{\mathbf{x}}_i||^2$ and $M = W_2W_1$. Then
\begin{align*}
 \mathbb{E}_{\epsilon} \left [SE(\tilde{\mathbf{x}}_i) \right ] & = \mathbb{E}_{\epsilon} \left [ || (I-M)\mathbf{x}_i + M(\mathbf{x}_i - \tilde{\mathbf{x}}_i) ||^2 \right ] \\
  & = SE(\mathbf{x}_i) + \mathbb{E}_{\epsilon} \left [ ||M(\mathbf{x}_i - \tilde{\mathbf{x}}_i)||^2 \right ]
\end{align*}
because the cross product terms vanish, since $\mathbb{E}_{\epsilon} \left [ \tilde{\mathbf{x}}_i \right ] = \mathbf{x}_i$: 
\begin{align*}
	0 & = \mathbb{E}_{\epsilon} \left [\mathbf{x}_i^T(I-M)^TM(\mathbf{x}_i - \tilde{\mathbf{x}}_i) \right ] \\
	& = \mathbb{E}_{\epsilon} \left [(\mathbf{x}_i - \tilde{\mathbf{x}}_i)^TM^T(I-M)\mathbf{x}_i \right ]. 
\end{align*}
We also have that
\begin{align*}
	||M(\mathbf{x}_i - \tilde{\mathbf{x}}_i)||^2 & = (\mathbf{x}_i - \tilde{\mathbf{x}}_i)^TM^TM(\mathbf{x}_i - \tilde{\mathbf{x}}_i) \\
	& = \textrm{tr}\left [ (\mathbf{x}_i - \tilde{\mathbf{x}}_i)^TM^TM(\mathbf{x}_i - \tilde{\mathbf{x}}_i) \right ] \\
	& = \textrm{tr}\left [ M(\mathbf{x}_i - \tilde{\mathbf{x}}_i)(\mathbf{x}_i - \tilde{\mathbf{x}}_i)^TM^T \right ] \\
	& = \textrm{tr}\left [ M\epsilon_i\epsilon_i^TM^T \right ]
\end{align*}
due to the invariance of the trace under cycle permutation of products. Therefore, in expectation over the noise we have
\begin{align*}
	\mathbb{E}_{\epsilon} \left [||M(\mathbf{x}_i - \tilde{\mathbf{x}}_i)||^2 \right ] = \textrm{tr}\left [ M(s^2 I)M^T \right ],
\end{align*}
and as a result
\begin{align*}
   \mathbb{E}_{\epsilon}\left [\mathcal{L} \right ] & = \frac{1}{2N} \sum^N_{i=1} ||\mathbf{x}_i - W_2W_1\mathbf{x}_i||^2 \nonumber \\
  & \textcolor{white}{white ce} + \frac{s^2}{2}\textrm{tr}\left ( W_2W_1W_1^TW_2^T \right).
\end{align*}

\subsection*{B. Learning dynamics for linear DAEs}

We derive the expression for the learning dynamics of a linear DAE as presented in (5) in the paper. As departure point, we start by examining the expected scalar update equations over the noise model for a small learning rate $\alpha$, which can be written as
\begin{align*}
    \tau \frac{d}{dt}w_1 & = w_2(\lambda - w_2w_1\lambda) - \varepsilon w_2^2w_1 \\
    \tau \frac{d}{dt}w_2 & = w_1(\lambda - w_2w_1\lambda) - \varepsilon w_2w_1^2.
\end{align*}
where $\tau = \frac{N}{\alpha}$, with $N$ representing the number of training samples. Define $w = w_2w_1$ and using the product rule the update for $w$ then becomes
\begin{align}
\tau \frac{d}{dt}w & = \tau [w_1\frac{d}{dt}w_2 + w_2\frac{d}{dt}w_1] \nonumber \\
& = w_1^2(\lambda - w_2w_1(\lambda + \varepsilon)) + w_2^2(\lambda - w_2w_1(\lambda + \varepsilon)) \nonumber \\
& = (\lambda - w(\lambda + \varepsilon))(w_1^2 + w_2^2). 
\label{eq: sup mat update for w}
\end{align}
Next we make the following hyperbolic change of coordinates
\begin{align*}
    w_1 = \sqrt{c_0}\text{sinh}\left(\frac{\theta}{2}\right), w_2 = \sqrt{c_0}\text{cosh}\left(\frac{\theta}{2}\right), \text{ for } w_1^2 < w_2^2 \\
    w_1 = \sqrt{c_0}\text{cosh}\left(\frac{\theta}{2}\right), w_2 = \sqrt{c_0}\text{sinh}\left(\frac{\theta}{2}\right), \text{ for } w_1^2 > w_2^2,
\end{align*}
where $\theta$ parameterises points along the dynamics trajectory represented by $w_2^2 - w_1^2 = \pm c_0$ \cite{saxe2013exact}. Note that with this change of coordinates we obtain
\begin{align*}
    w & = c_0\text{cosh}\left( \frac{\theta}{2} \right)\text{sinh}\left( \frac{\theta}{2} \right)\\
    & = c_0 \left ( \frac{e^{\frac{\theta}{2}} + e^{-\frac{\theta}{2}}}{2} \right ) \left ( \frac{e^{\frac{\theta}{2}} - e^{-\frac{\theta}{2}}}{2} \right ) \\
    & = \frac{c_0}{2}\left ( \frac{e^{\theta} - e^{-\theta}}{2} \right ) \\
    & = \frac{c_0}{2}\text{sinh}(\theta),
\end{align*}
so that 
\begin{align*}
    dw = \frac{c_0}{2}\text{cosh}(\theta)d\theta.
\end{align*}
Similarly,
\begin{align*}
    w_2^2 + w_1^2 & = c_0\text{cosh}^2\left( \frac{\theta}{2} \right) + c_0\text{sinh}^2\left( \frac{\theta}{2} \right) \\
    & = c_0\left ( \frac{e^{\frac{\theta}{2}} + e^{-\frac{\theta}{2}}}{2} \right )^2 + c_0\left ( \frac{e^{\frac{\theta}{2}} - e^{-\frac{\theta}{2}}}{2} \right )^2 \\
    & = \frac{c_0}{4} \left ( e^{\theta} + 2 + e^{-\theta} + e^{\theta} -2 + e^{-\theta} \right )\\
    & = c_0\left ( \frac{e^{\theta} + e^{-\theta}}{2} \right ) \\
    & = c_0\text{cosh}(\theta)
\end{align*}
Plugging these results into the update for $w$ given in \eqref{eq: sup mat update for w}, yields
\begin{align*}
    \frac{\tau c_0\text{cosh}(\theta)}{2} \frac{d\theta}{dt} = \left (\lambda - \frac{c_0}{2}\text{sinh}(\theta)(\lambda + \varepsilon) \right )c_0\text{cosh}(\theta),
\end{align*}
and as a result,
\begin{align*}
    \tau \frac{d\theta}{dt} = \lambda \left (2 - \beta\text{sinh}(\theta)\right ),
\end{align*}
where $\beta = c_0\left(1 + \frac{\varepsilon}{\lambda}\right)$. To solve for $t$, we write
\begin{align*}
    t & = \int^{\theta_f}_{\theta_0} \frac{\tau}{\lambda\left ( 2 - \beta\text{sinh}(\theta) \right )} d\theta 
\end{align*}
and integrate:
\begin{align*}
    t & = \frac{\tau}{\zeta\lambda}\left [ \text{ln} \left ( \frac{\zeta + \beta + 2\text{tanh}(\frac{\theta}{2}) }{\zeta - \beta - 2\text{tanh}(\frac{\theta}{2})}\right )\right ]^{\theta_f}_{\theta_0}
\end{align*}
where $\zeta = \sqrt{\beta^2 + 4}$ and initial parameter value $\theta_0 = \text{sinh}^{-1}(2w/c_0)$. Let $\delta_0 = \text{tanh}\left(\frac{\theta_0}{2}\right)$ and $\delta_f = \text{tanh}\left(\frac{\theta_f}{2}\right)$,  then 
\begin{align*}
    t & = \frac{\tau}{\lambda \zeta} \text{ln} \frac{\left ( \zeta + \beta + 2\delta_f \right ) \left ( \zeta - \beta - 2\delta_0 \right )}{\left ( \zeta - \beta - 2\delta_f \right ) \left ( \zeta + \beta + 2\delta_0 \right )} ,
\end{align*}
so that
\begin{align*}
    e^{\lambda \zeta t / \tau} & = \frac{\left ( \zeta + \beta + 2\delta_f \right ) \left ( \zeta - \beta - 2\delta_0 \right )}{\left ( \zeta - \beta - 2\delta_f \right ) \left ( \zeta + \beta + 2\delta_0 \right )} .
\end{align*}

Multiplying by the denominator, expanding, and defining $E = e^{\lambda \zeta t / \tau}$, we obtain
\begin{align*}
    & -2 E \delta_f \left (\zeta + \beta + 2 \delta_0 \right ) \\
    & + E \left (\zeta^2 + 2 \zeta \delta_0 - \beta^2 - 2 \beta \delta_0 \right ) \\
    & = 2 \delta_f \left ( \zeta - \beta - 2 \delta_0 \right ) \\
    & + \left ( \zeta^2 - 2 \zeta \delta_0 - \beta^2 - 2 \beta \delta_0 \right ) ,
\end{align*}
which yields
\begin{align*}
    & \delta_f \left ( (1 - E) \left ( 2 \beta + 4 \delta_0 \right ) - 2 (E + 1) \zeta \right ) \\
    & = (1 - E) \left ( \zeta^2  - \beta^2 - 2 \beta \delta_0 \right ) - 2 (1 + E) \zeta \delta_0 .
\end{align*}

Solving for $\theta_f(t)$, we obtain the hyperbolic parameter equation
\begin{align*}
    \theta_f(t) = 2\text{tanh}^{-1}\left [ \frac{(1-E)\left (\zeta^2 - \beta^2 - 2\beta\delta \right) - 2(1 + E)\zeta \delta}{(1-E)\left ( 2\beta + 4\delta\right) - 2(1+E)\zeta} \right ]
\end{align*}
where $\delta = \text{tanh}\left(\frac{\theta_0}{2}\right)$. Using 
\begin{align*}
w(t) = \frac{c_0}{2}\text{sinh}\left (\theta_t \right),
\end{align*}
(where $\theta_t = \theta_f(t)$) to track the weight trajectory gives equation (5) in the paper.

\subsection*{C. Learning dynamics for linear WDAEs}

We derive the expression for the learning dynamics of a linear WDAE as presented in (7) in the paper. Reconstruction loss with weight decay gives the scalar loss associated with an eigenvalue $\lambda$ as
\begin{align*}
\ell_{\gamma} = \frac{\lambda}{2\tau}(1 - w_2w_1)^2 + \frac{N\gamma}{2\tau}(w_1^2 + w_2^2),
\end{align*}
where $\gamma$ is the penalty parameter that controls the amount of regularisation that is being applied. The update equations for the weights then follow as
\begin{align*}
    \tau \frac{d}{dt}w_1 & = w_2(\lambda - w_2w_1\lambda) - N\gamma w_1 \\
    \tau \frac{d}{dt}w_2 & = w_1(\lambda - w_2w_1\lambda) - N\gamma w_2,
\end{align*}
assuming the initial $w_2 = w_1$ (which holds approximately for small initial values), we have for $w = w_2w_1$ that
\begin{align*}
    \tau \frac{d}{dt}w & = 2w(\lambda - w\lambda) - 2N\gamma w \\
    & = 2w(\lambda - N\gamma - w\lambda).
\end{align*}
Thus, 
\begin{align*}
    t & = \int^{w_f}_{w_0} \frac{\tau}{2w(\lambda - N\gamma - w\lambda)}dw \\
    & = \frac{\tau}{2}\left [ \frac{\text{ln}(w) - \text{ln}(\lambda-N\gamma-w\lambda)}{\lambda-N\gamma} \right ]^{w_f}_{w_0} \\
    & = \frac{\tau}{2(\lambda - N\gamma)}\text{ln} \left ( \frac{w_f(\lambda-N\gamma - w_0\lambda)}{w_0(\lambda-N\gamma - w_f\lambda)} \right ).
\end{align*}
Then solving for $w_f$ gives 
\begin{align*}
w_f(t) = \frac{\xi E_\gamma}{E_\gamma-1 + \xi/w_0},
\end{align*}
where $E_\gamma = e^{2\xi t /\tau}$ and $\xi = (1 - N\gamma/\lambda)$.

\subsection*{D. Optimal learning rates}

We derive expressions for the optimal learning rates for linear DAEs and WDAEs as presented in (8) in the paper. First, consider the expected scalar DAE loss
\begin{align*}
	\ell_\varepsilon = \frac{\lambda}{2\tau}(1 - w_2w_1)^2 + \frac{\varepsilon}{2\tau}(w_2w_1)^2.
\end{align*}
The Hessian of $\ell_\varepsilon$ is given by
\begin{align*}
	H = \begin{bmatrix}
    \frac{\partial^2 \ell_\varepsilon}{\partial w_1^2} & \frac{\partial^2 \ell_\varepsilon}{\partial w_1w_2} \\
    \frac{\partial^2 \ell_\varepsilon}{\partial w_2w_1} & \frac{\partial^2 \ell_\varepsilon}{\partial w_2^2} \\
\end{bmatrix},
\end{align*}
where 
\begin{align*}
	\frac{\partial^2 \ell_\varepsilon}{\partial w_1^2} & = \frac{w_2^2}{\tau}(\lambda + \varepsilon), \\
	\frac{\partial^2 \ell_\varepsilon}{\partial w_2^2} & = \frac{w_1^2}{\tau}(\lambda + \varepsilon), \\
	\frac{\partial^2 \ell_\varepsilon}{\partial w_1w_2} & = \frac{\partial^2 \ell_\varepsilon}{\partial w_2w_1} = \frac{2w_2w_1}{\tau}(\lambda + \varepsilon) - \frac{\lambda}{\tau}.
\end{align*}
Now, if we assume $w_2 = w_1$, and let $a = \frac{\partial^2 \ell_\varepsilon}{\partial w_1^2} = \frac{\partial^2 \ell_\varepsilon}{\partial w_2^2}$ and $b = \frac{\partial^2 \ell_\varepsilon}{\partial w_2w_1}$, the eigenvalues for the Hessian can be shown to be
$\lambda_H = a-b$ or $\lambda_H = a + b$. 
The second order update for a single weight $w$ at time $t$ is then given by 
\begin{align*}
	w^{t+1} = w^t - \left(\frac{\partial \ell_\varepsilon}{\partial w^t}\right)/\lambda_H,
\end{align*}
where the maximum $\lambda_H$, is when $w_2 = w_1 = 1$, such that 
\begin{align*}
	\lambda_H & = \frac{1}{\tau}(\lambda + \varepsilon) + \frac{2}{\tau}(\lambda + \varepsilon) - \frac{\lambda}{\tau} \\
	& = \frac{2\lambda + 3\varepsilon}{\tau}.
\end{align*}
Therefore, the optimal learning rate is
\begin{align*}
	\alpha_\varepsilon = 1/\lambda_H = \frac{\tau}{2\lambda + 3\varepsilon}.
\end{align*}
For WDAEs with penalty parameter $\gamma$, a very similar derivation gives
\begin{align*}
	\alpha_\gamma = \frac{\tau}{2\lambda + \gamma}.
\end{align*}
Taking the ratio of the optimal DAE rate to that for the WDAE gives
\begin{align*}
	R = \frac{\alpha_\varepsilon}{\alpha_\gamma} = \frac{2\lambda + \gamma}{2\lambda + 3\varepsilon}.
\end{align*}

\subsection*{E. Equivalent scalar solutions}

In Section 4 of the paper, the DAE fixed point solution is shown to be
\begin{align*}
	w^*_\varepsilon = \frac{\lambda}{\lambda + \varepsilon}.
\end{align*}
Now if $w = w_2w_1$ and $w_2 = w_1$, then for WDAE we have that the scalar loss is given by
\begin{align*}
 \ell_{\gamma} = \frac{\lambda}{2\tau}(1 - w)^2 + \frac{\gamma}{\tau}w,
\end{align*}
and 
\begin{align*}
\frac{\partial \ell_{\gamma}}{\partial w} = -\frac{\lambda}{\tau}(1 - w) + \frac{\gamma}{\tau}.
\end{align*}
Setting the above equal to zero and solving gives
\begin{align*}
	w^*_\gamma = 1-\gamma/\lambda.
\end{align*}
To obtain the value of $\gamma$ for which the two fixed points are equal, we set $w^*_\gamma = w^*_\varepsilon$ and solve for $\gamma$ to find
\begin{align*}
	\gamma = \frac{\lambda \varepsilon}{\lambda + \varepsilon}.
\end{align*}

\subsection*{F. Estimated dynamics for nonlinear networks}

The dynamics for the nonlinear networks trained in Figure 6 in the paper were estimated using the following approach. First, compute 
\begin{align*}
	\Sigma_{xx} = \sum^N_{i=1}\mathbf{x}_i\mathbf{x}_i^T = V \Lambda V^T,
\end{align*}
using an eigen-decomposition giving eigenvalues $\lambda_j, j = 1, ..., D$. Then at regular intervals compute 
\begin{align*}
	\hat{\Sigma}_{xx}(t) = \sum^N_{i=1}\mathbf{x}_i\hat{\mathbf{x}}_i(t)^T,
\end{align*}
where $\hat{\mathbf{x}}(t)$ is the estimated reconstruction of input at time $t$ generated by the autoencoder network. Finally, using the following rotation to obtain the diagonal matrix 
\begin{align*}
	\hat{\Lambda}(t) = V^T\hat{\Sigma}_{xx}(t)V,
\end{align*}
where the diagonal contains the estimated eigenvalues $\hat{\lambda}_j(t)$, we can compute an estimate for the identity mapping associated with each eigenvalue as $\hat{\lambda}_j(t)/\lambda_j \in [0, 1]$.

\subsection*{G. Learning dynamics for tanh autoencoder networks}

We investigated the dynamics of learning for nonlinear AEs, WDAEs and DAEs, using tanh activations. 

\begin{figure}[h]
   \centering
  \includegraphics[width=0.99\linewidth]{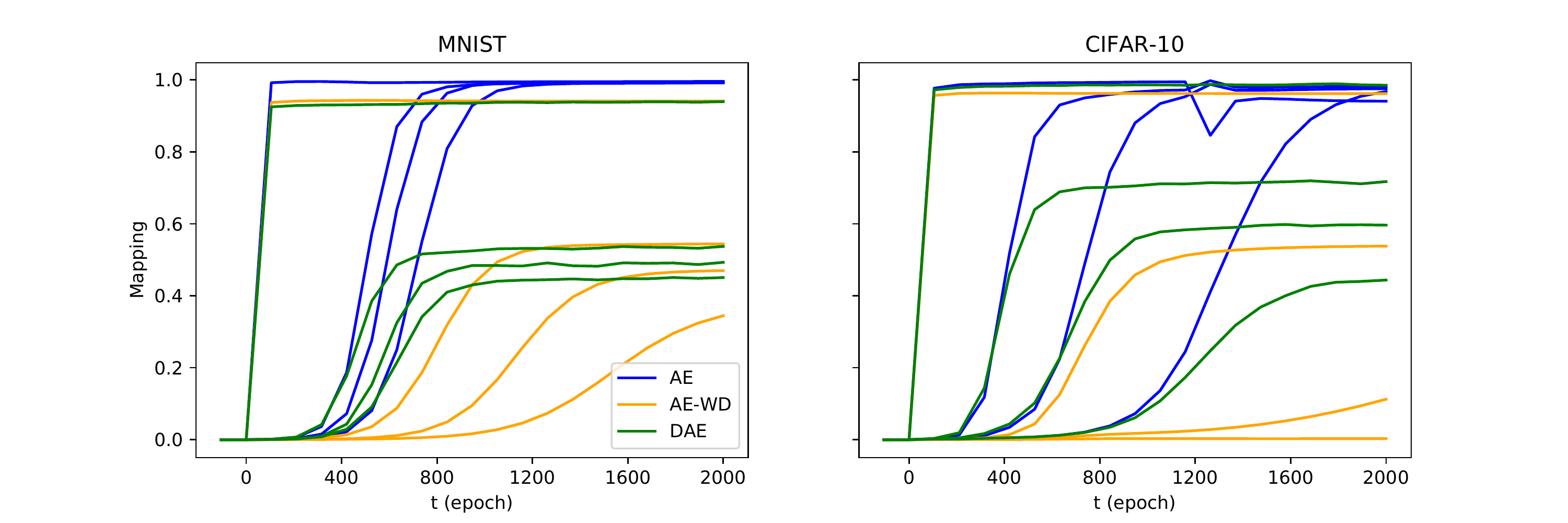} 
   \caption{\textit{Learning dynamics for nonlinear networks using tanh activation.} AE (blue), WDAE (orange) and DAE (green). \textbf{Left}: MNIST \textbf{Right}: CIFAR-10.}
   \label{fig: nonlinear tanh learning dynamics}
\end{figure}

Figure \ref{fig: nonlinear tanh learning dynamics} shows the dynamics for these networks trained on MNIST ($N = 50000$) and CIFAR-10 $(N=30000)$ with equal learning rates. For the DAE, the input was corrupted using sampled Gaussian noise with mean zero and $\sigma^2 = 2$. For the WDAE, the amount of weight decay was set to $\gamma = 0.0045$. During the course of training, the identity mapping associated with each eigenvalue was estimated using the approach described in Section F, at equally spaced intervals of size $100$ epochs.

\end{document}